\documentclass[letterpaper, 10 pt, conference]{IEEEtran} 

\IEEEoverridecommandlockouts                             
\usepackage[utf8]{inputenc}
\usepackage[T1]{fontenc}
\usepackage{graphicx}
\usepackage{subcaption}

\usepackage{calc}

\usepackage{amssymb}
\usepackage{amstext}
\usepackage{amsmath}
\usepackage{amsthm}
\usepackage{multicol}

\title{\LARGE \bf
3D Object Detection From LiDAR Data Using Distance Dependent Feature Extraction
}

\author{Guus Engels$^{1}$, Nerea Aranjuelo$^{1}$, Ignacio Arganda-Carreras$^{2}$, Marcos Nieto$^{1}$ \& Oihana Otaegui$^{1}$% <-this % stops a space
\thanks{$^{1}$Guus Engels, Nerea Aranjuelo, Marcos Nieto \& Oihana Otaegui are with the Vicomtech Foundation, Basque Research and Technology Alliance (BRTA), Mikeletegi 57, San Sebastian, Spain. 
        {\tt\small \{gengels, naranjuelo, mnieto, ootaegui\}@vicomtech.org}}%
        
\thanks{$^{2}$Ignacio Arganda-Carreras is with 1.) Basque Country University (UPV/EHU), San Sebastian, Spain, 2.) Ikerbasque, Basque Foundation for Science, Bilbao Spain \& 3.) Donostia International Physics Center (DIPC), San Sebastian, Spain.
        {\tt\small ignacio.arganda@ehu.eus}}
}

\begin{document}

\maketitle
\thispagestyle{empty}
\pagestyle{empty}

%%%%%%%%%%%%%%%%%%%%%%%%%%%%%%%%%%%%%%%%%%%%%%%%%%%%%%%%%%%%%%%%%%%%%%%%%%%%%%%%
\begin{abstract}

This paper presents a new approach to 3D object detection that leverages the properties of the data obtained by a LiDAR sensor.
State-of-the-art detectors use neural network architectures based on assumptions valid for camera images. However, point clouds obtained from LiDAR data are fundamentally different.
Most detectors use shared filter kernels to extract features which do not take into account the range dependent nature of the point cloud features. 
To show this, different detectors are trained on two splits of the KITTI dataset: close range (points up to 25 meters from LiDAR) and long-range. Top view images are generated from point clouds as input for the networks. Combined results outperform the baseline network trained on the full dataset with a single backbone.
Additional research compares the effect of using different input features when converting the point cloud to image. The results indicate that the network focuses on the shape and structure of the objects, rather than exact values of the input.
This work proposes an improvement for 3D object detectors by taking into account that features change over distance in point cloud data. Results show that training separate networks for close-range and long-range objects boosts performance for all KITTI benchmark difficulties.

\end{abstract}

%%%%%%%%%%%%%%%%%%%%%%%%%%%%%%%%%%%%%%%%%%%%%%%%%%%%%%%%%%%%%%%%%%%%%%%%%%%%%%%%
\section{INTRODUCTION}

\noindent "LiDAR is a fool's errand and anyone relying on LiDAR is doomed." - Elon Musk, CEO of Tesla.
The CEO of Tesla, a company that puts tremendous effort in the development of autonomous vehicles does not see value in using the LiDAR sensor, but a vast number of researchers and companies disagree and have shown that including LiDAR in their perception pipeline can be beneficial for advanced scene understanding.

\begin{figure}[ht!]
\begin{subfigure}[t]{0.49\linewidth}
\includegraphics[width=\linewidth, height=3.2cm]{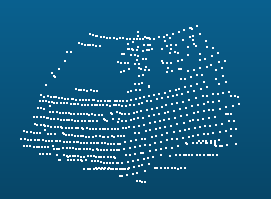}
\end{subfigure}
\begin{subfigure}[t]{0.485\linewidth}
\includegraphics[width=\linewidth, height=3.2cm]{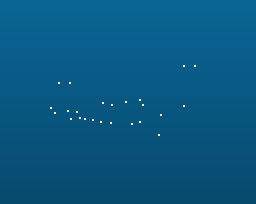}
\end{subfigure}\\
\begin{subfigure}[t]{0.99\linewidth}
\includegraphics[width=\linewidth]{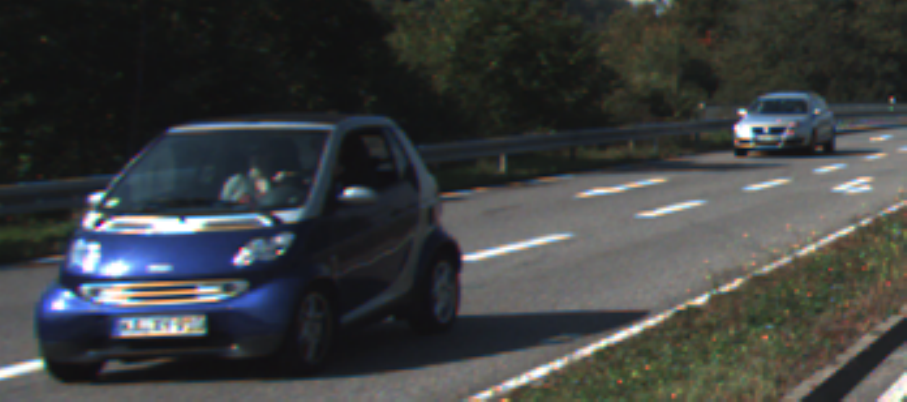}
\end{subfigure} \\
\caption{Frame from the KITTI dataset with the camera image on the bottom and two point cloud crops from the cars on the top figures. The left car on the image and the top left point cloud representation belong to each other and the top right and right car.
}
\label{firstim}
\end{figure}

LiDAR is an abbreviation for light detection and ranging. It is a sensor that uses laser light to measure the distance to objects and their reflectiveness. It sends out beams of light that diverge over the distance and reflect if an object is hit. It has proven to be useful for many applications in areas such as archaeology and geology. However, there are some downsides that prohibit the adoption of widespread use for autonomous driving.
Currently, LiDAR is an expensive, computationally demanding sensor that requires some revision of the car's layout. Cameras are a rich source of information that are very cheap in comparison.
Nonetheless, there are situations where cameras struggle and LiDAR can be of significant help.
Driving in low light conditions is challenging for a camera, but does not make a difference for LiDAR because it does not depend on external light sources.
Overall, the future of LiDAR and its role in a future autonomous vehicle is unclear, but it does have very valuable properties that can play an important role in autonomous driving and make its research decisive.

For self-driving vehicles it is of the utmost importance to be aware of its surroundings. Therefore, objects have to be located in the 3D space around the ego vehicle, which requires using sensors such as LiDAR to capture the surrounding information. Research into 3D object detection and LiDAR have gone hand in hand.
The most influential dataset for 3D object detection for autonomous applications has been the KITTI dataset \cite{kitti}. It hosts leaderboards where different methods are compared. The growing interest in 3D object detection can be observed from these leaderboards, where the state of the art gets replaced quickly by newer architectures.
The best performing network in 2018, Pixor \cite{Pixor}, only just falls in the top 100 as of October 2019.
All detectors on the leaderboards, used for 3D object detection are influenced in varying degrees by 2D detectors. It is obvious that the advances in 2D object detection can be of big help for 3D object detection. However, 2D object detection is done on images, a very different data type compared to point cloud data. The LiDAR beams diverge, which causes objects placed farther away to be represented with less points than the exact same object nearby. This is very different compared to images, where the object becomes smaller when it is farther away, but the underlying features and its representation are still the same. This can be clearly seen from Figure \ref{firstim}.
The effect of this has not yet received much attention in the literature.

In this paper, differences between image and point cloud data are researched. A new pipeline is proposed that learns different feature extractors for objects that are close by and far away. This is necessary because features in these two ranges are very different, and one shared feature extractor would perform sub-optimally in both ranges.
Furthermore, visualization of the changing features over the distance is provided. Previous works like ZFnet \cite{zfnet} were not only able to increase performance, but also showed the importance of understanding how the network operates. Inspired by this we focus on how point cloud input data is different from image input data to have a better understanding of the limitations of using 2D CNNs on point clouds. In addition we show how to alter the detection pipeline to exploit these differences. 

To summarize, the contributions of this paper are the following:
\begin{itemize}
    \item We provide an analysis into the effects of different input features on the detection performance.
    \item We show the influence of the changing representation over distance and how it affects the underlying features of the objects on the car class.
    \item We propose a new detection pipeline that is able to learn different feature extractors for close-by objects and far-away objects.
\end{itemize}

\section{RELATED WORK}

\noindent 3D object detection has been tackled with many different approaches. The first networks were heavily influenced by 2D detectors. They converted point clouds to top view images, which made it possible to use conventional 2D detectors. A variety of input features and conversion methods were used.
Many of these detectors have been later on improved by adding more information from other sensors or maps.

Currently, most of the best performing architectures use methods that directly process the point cloud data \cite{partAware,sparse2dense}. They do not require handcrafted features or conversion to images like the previous methods did. As of late, new datasets have become public and will push the field of 3D object detection further forward.

\subsection{2D CNN approaches}

\noindent Many early and current 3D object detectors are adjusted 2D object detectors such as Faster R-CNN \cite{fasterrcnn} and YOLO \cite{Yolo}. In particular, ResNet \cite{resnet} is often used as backbone network for feature extraction. In order to detect 3D objects with 2D detectors, point clouds are compressed to image data. Two common representations are (i) a front view approach like in LaserNet \cite{lasernet} and (ii) a bird's eye view (BEV) approach where point cloud data are compressed in the height dimension \cite{birdnet} . Some methods only use three input channels to keep the input exactly the same as for 2D detectors. Other methods use more height channels to save more important height information of the point cloud \cite{Pixor,m3vd}.
Which features are important to retain has been researched before but is still an open research topic. The BEV has proven to be such a powerful representation that methods using camera data convert their data to a BEV to perform well on 3D object detection tasks \cite{pseudolidar}. How the conversion from point cloud to BEV is performed will be explained in the Methods section.

\subsection{Multimodal approaches}

A modern autonomous vehicle will have more than just a LiDAR sensor. This is why many methods use a multimodal approach that combines LiDAR with camera sensors \cite{m3vd,multisensor,ku2018joint,frustumpointnet}. Other methods fuse the data at different stages in the network. MTMS \cite{multisensor} uses an earlier fusion method, whereas M3VD \cite{m3vd} uses a deep fusion method.
Besides different sensors, map information can also be employed to improve performance \cite{HDNET}. Some approaches track objects over frames \cite{fastandfurious,3dmultiobj} which boosts performance, especially for objects with only a small number of LiDAR points and partially occluded objects in a sequence.

\subsection{Detection directly on the point cloud}

A major shift in 3D object detection came when networks started to extract features directly from the point cloud data. Voxelnet \cite{voxelnet} introduced the VFE layer, which computed 3D features from a set of points in the point cloud and stores the extracted value in a voxel, containing the value of a volume in 3D space, similar to what a pixel does for 2D space. 
A second influential network that works directly with the point cloud data is PointNet \cite{PointNet}. It processes directly on the unordered point cloud and is invariant to transformations, such as translations and rotations. These two methods are the foundation of many architectures in the top 50 on the KITTI leaderboards. Pointpillars \cite{pointpillars} uses PointNet to extract features and then transform it to a BEV representation to apply a detector with a FPN inspired backbone \cite{FPN}. Other networks use this approach and focus on the orientation \cite{second} or try bottom up anchor generation \cite{PointRCNN,IPOD}.

\begin{figure*}
\includegraphics[width=\textwidth]{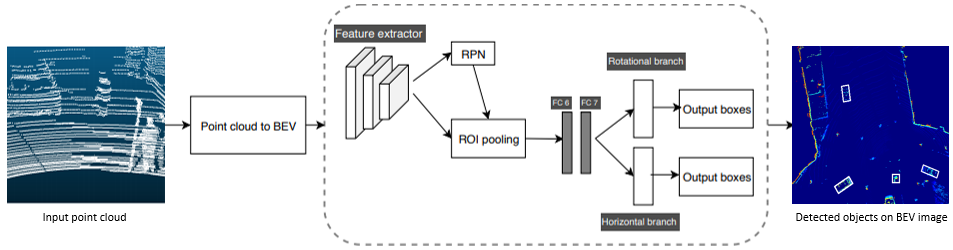}
\caption{Proposed pipeline for 3D object detection using point clouds as input. In this method, point clouds are converted to bird's eye view (BEV) images in the "point cloud to BEV" stage. The point cloud is compressed to an image according to one of four configurations explained in the Methods section. The architecture inside the dotted lines is a Faster R-CNN architecture with an additional rotation branch that makes it possible to output rotated bounding boxes. RPN stands for region proposal network, which searches for regions of interest for the head-network that tries to classify and regress bounding boxes around these objects. ROI pooling converts different sized regions of interest to a common $7\times7$ so it can be handled by two fully connected layers called FC6 and FC7. The output of these layers are send to the last branches that predict either rotated or horizontal bounding boxes with a class label.
}
\label{pipeline}
\end{figure*}

\subsection{Datasets}

The KITTI dataset \cite{kitti} has been one of the most widely used datasets for 3D object detection for autonomous driving so far. It is a large scale dataset that contains annotations in the front camera field of view for camera and LiDAR data. There are three main classes, namely, cars, pedestrians and cyclists. However, the car class contains more than half of all objects \cite{kitti}.
As of late, many new datasets have emerged. Most notable are NuScenes \cite{Nuscenes} and the H3D dataset, of Honda \cite{h3ddataset}. They are much larger datasets, 360 degrees annotated and contain 1 million and 1.4 million bounding boxes (respectively) compared to 200k annotations in KITTI.
The NuScenes dataset was created with a 32-channel LiDAR, in comparison to the KITTI and H3D dataset which were both created with a 64-channel LiDAR. Sensors with more channels are able to produce denser, more complete 3D maps. The beam divergence will be larger when there are less channels and will make it more difficult to detect objects far away.

\section{METHODS}

\noindent
In order to apply 3D object detection on LiDAR and perform the different tests mentioned in the Introduction, a 3D object detection algorithm is implemented according to the structure in Figure \ref{pipeline}. The point cloud is converted to images, which can then be used by a Faster R-CNN detection network. This network is used to evaluate the effect of different input features on the performance and to create an architecture that uses different kernel weights for different regions of the LiDAR point cloud.

\subsection{Network}

\subsubsection{Architecture}
The architecture of the full baseline network is displayed in Figure \ref{pipeline}. The full architecture has a point cloud as input and outputs horizontal and rotational bounding boxes. 
A Faster R-CNN \cite{fasterrcnn} with a rotation regression branch in the head network is used. The backbone network is ResNet-50 \cite{resnet} of which only the first three blocks are used for the RPN network. Faster R-CNN is designed for camera images where objects can have all kinds of shapes. In a BEV image of LiDAR data all objects are relatively small in comparison. To make sure that the feature map is not downsampled in size too quickly, the stride of the first ResNet block is adjusted from 2 to 1. This causes the output feature map to be four times as big, which means that less spatial information is lost.
The object size in a BEV image is directly proportional to the physical size of that object. This allows for tailoring the anchor boxes exactly to the object size to serve as good priors. When cars are rotated these anchor boxes will not fit accurately to the objects. To account for this, different orientations of the anchor boxes are considered similarly to \cite{r2cnn_anchor,complexer-yolo}. In total, sixteen different orientations in the anchor boxes are used.
The head network is slightly different from Faster-RCNN because it contains an additional rotation head \cite{rotation}. 
Region proposals from the RPN network and the output feature map of the RPN network are the inputs of the head network. Region of interest pooling (ROI) outputs $7\times7\times1024$ feature maps for each object proposal that are fed to two fully connected layers with $1024$ parameters. These layers output four bounding box variables for non rotated objects, more specifically, the center of the box coordinates ($x$, $y$) and the dimensions ($h$, $w$). The rotated boxes have an additional parameter, $\theta$, for the rotation angle of the bounding box. Each box has a class confidence score. The class with the highest score is the most likely for this bounding box and it is used as output.

\subsubsection{Point cloud to BEV Image}

\begin{figure}
\includegraphics[width=0.47\textwidth,height=6
cm]{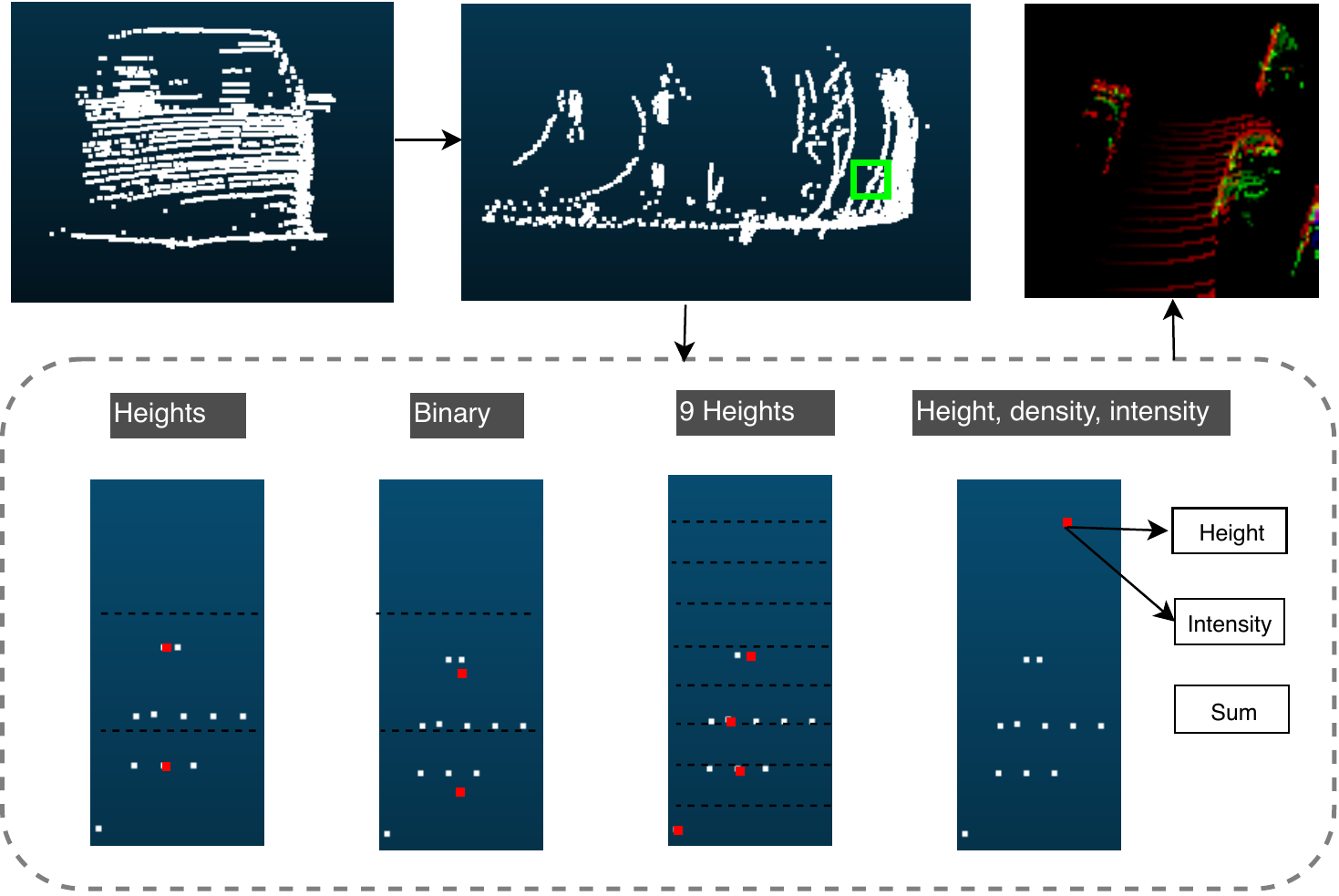}
\caption{Point cloud conversion to bird's eye view (BEV) representation. This research uses four different possible configurations that are visualized in the bottom dotted block. In the complete pipeline only one of these will be used for each test.}
\label{bevconversion}
\end{figure}

Feature extractors such as ResNet are designed for RGB images. Using this exact architecture requires reshaping the point cloud input to a three channel image.
A common approach consists of compressing the point cloud into a BEV image where the height dimension is represented by three channels. Which information to keep remains an open problem. In the past, it was shown that the reflectiveness, or intensity value, did not contribute much when there was maximum height information available \cite{birdnet}. Pixor \cite{Pixor} and MV3D \cite{m3vd} use multiple height maps to retain more height information. Instead of only storing 3 channels, as it would be the case in a regular 2D detector, they have an architecture that can handle inputs of more than 3 channels.
To test the effect of different features, the inputs are pre-processed in four different input configurations. For all methods there are some shared steps. Firstly, a 3D space of $0$m to $70$m in longitudinal range, $-35$m to $35$m in lateral range and $-1.73$m to $1.27$m in the height dimension is defined. The LiDAR in the KITTI setup is located $1.73$m off the ground, so $-1.73$m is where the road is. All points that fall outside of this box are not taken into account. For three of the four methods, $700\times700\times3$ pixel images are considered, which means that every $0.1\times0.1\times3$m cube represents one pixel. The last method uses more channels and is of size $700\times700\times9$. The exact specifications of the channels are as follows:

\begin{figure*}
\centering
\includegraphics[width=\textwidth,height=4cm]{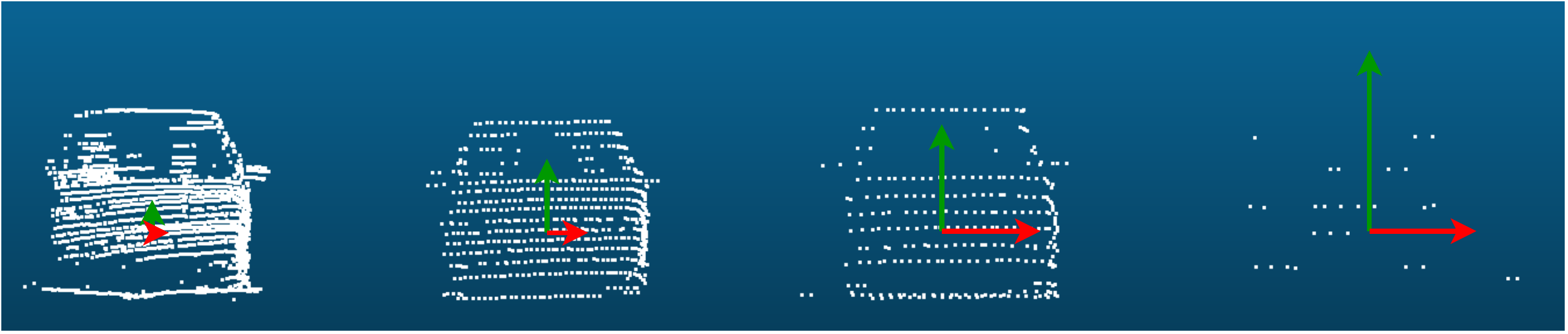}
\caption{Representation of four cars at different ranges detected with a 64 channel LiDAR. The green arrow indicates the vertical distance between four adjacent points and the red arrow indicates the horizontal distance between four adjacent points. The distance of four points instead of the distance between adjacent points is used for visualization purposes.}

\label{pointcloudcars}
\captionof{table}{Statistics of the 4 cars, ordered from left to right.}
\begin{tabular}{l|llll}
\textbf{Car number}       & 1    & 2    & 3     & 4     \\ \hline
\textbf{distance} [m]        & 7.6   & 16.0   & 24.4    & 43.0   \\
\textbf{vert spacing} [cm]     & 4 & 9  & 13 & 22  \\
\textbf{hor spacing}  [cm]    & 2  & 5  & 8  & 13  \\
\textbf{amount of points} & 1797 & 627  & 226   & 25   
\end{tabular}

\end{figure*}

\textbf{Max height voxels}: the cube can be divided in three voxels with each $0.1$m $\times$ $0.1$m $\times$ $1$m dimensions. The first channel will have the height value of the highest point from  $-1.73$m to $-0.73$m range. The second channel will have the highest value from $-0.73$m to $0.23$m and the third one from $0.23$m to $1.23$m. An advantage of this method is that automatically a form of ground removal takes place. The points very close to the road are often not of much importance for the objects on the road. In addition, the highest points of an object are in many cases what distinguishes them, and therefore often the most important feature \cite{birdnet}. 

\textbf{Binary}: for every voxel it is checked if there is at least one LiDAR point present. If this is the case, this channel gets the value $100$, if not it gets the value $0$. The value $100$ is chosen instead of $1$ to make sure the values have the same order of magnitude as the pre-trained Imagenet \cite{imagenet} weigths. This approach is an important indicator of how the network handles the specific point cloud values. Consequently, it is a baseline test to see if other features add value. 

\textbf{Multichannel max height voxels}: the input of only three channels might not be enough since it means that many of the original point cloud information is lost. Using more height will provide the network with more of the original information and should improve the scores as was done in \cite{m3vd,Pixor}. Nine height maps are used as input to the ResNet backbone instead of the three maps used in the max height approach. 
To make sure that this input is compatible with the ResNet feature extractor, the pre-trained weights are duplicated three times and stacked.

\textbf{Height intensity density}: instead of picking a subset of the points from the point cloud, it is also possible to compute features that could be interesting and feed them to the network. This configuration has been used in Birdnet \cite{birdnet} before. 

A visualization of how all the features are extracted is displayed in Figure \ref{bevconversion}. This stage represents what happens in the "Pointcloud to BEV" block in Figure \ref{pipeline}.

\subsection{Loss function}
Faster R-CNN is a two-stage detector for which the total loss is a combination of the losses of the individual stages. The losses of the first stage are similar to the original Faster R-CNN paper, but do not have the logarithmic scaling factors. There are regression targets that describe the absolute difference between the ground truth and the network prediction for the center points ($x$, $y$) and the dimensions ($h$, $w$) of the object: 

\begin{align}
{\Delta}x_c&=x_c - x_{ct}  \\       
{\Delta}y_c&=y_c - y_{ct} \\
{\Delta}h&=h - h_{t} \\         
{\Delta}w&=w - w_{t}
\end{align}

where $x_c, y_c, h, w$ are predictions from the network and $x_{ct}, y_{ct}, h_{t}, w_{t}$ are the ground truth targets which the network aims to predict. 
The classification loss is a softmax cross-entropy between the background and foreground classes. Together they form the RPN loss.

The horizontal branch of the head network uses the exact same regression targets, but should now consider multiple, instead of just background and foreground classes. In this research, only the car class is considered so it turns out to be the same as for the RPN case.
The rotational branch of the head network has an additional regression loss for the rotation as show in Figure \ref{pipeline}. This loss is the absolute difference between the predicted and ground truth angle in degrees:
\begin{equation}
{\Delta}\theta=\theta - \theta_{t}
\end{equation}

It is important to note that by introducing rotation it becomes possible to describe the exact same rotated bounding box in multiple ways. Consider a bounding box defined by $x_c$, $y_c$, $h$, $w$, and $\theta$. If the values of $h$ and $w$ are swapped and $90$ degrees are added to the angle $\theta$, the same bounding box can be described. This has to be avoided since a loss larger than zero would be possible, even though the predict bounding box fits perfectly. This is avoided by making sure that the width is always larger than the height. If this is not the case the height and width are swapped and 90 degrees are added to the rotation to force unique bounding box configurations. 

Once all regression targets are calculated smooth L1 loss is used according to the following equation:

\begin{equation}
L1 (x) = {
\begin{cases} 
0.5x^2 \sigma\; \qquad \text{if} \ |x| < \frac{0.5}{\sigma}\\ 
|x| - \frac{0.5}{\sigma} \qquad \text{otherwise}
\end{cases}
}
\end{equation}

where $\sigma$ is a tuning hyperparameter, $\sigma=3$  for the RPN network and $\sigma=1$ for the head network are used, which is in line with Faster R-CNN implementation \cite{fasterrcnn}.

Combining the classification and regression losses for all branches results in six total losses. A classification and regression loss for the RPN, horizontal and rotational branch. All losses use smooth L1 loss to calculate the total loss value according to the following equation:

\begin{align}
\begin{split}
  Loss &= L{_{rpn,reg}} + L{_{rpn,class}} + L{_{head_h,reg}}  \\
  &+L{_{head_h,class}} +L{_{head_r,reg}} + L{_{head_r,class}}  \\
\end{split}
\end{align}

where $L_{rpn,reg}$ is the regression loss of the RPN, $L_{rpn,class}$ is the classification loss of the RPN, $L_{head_h,reg}$ is the regression loss of the horizontal head network, $L_{head_h,class}$ is the classification loss of the horizontal head network, $L_{head_r,reg}$ is the regression loss of the rotational head network and $L_{head_r,class}$ is the classification loss of the rotational head network.

\subsection{Working with LiDAR data}

The density and distance between points change over distance in the point clouds. Deep learning architectures are often designed with images in mind where this is not the case. The section shows how features change in the LiDAR data and how the network architecture can be adjusted to account for this.

\subsubsection{LiDAR features}
An important reason why convolutional neural networks are able to work well with relatively few parameters compared to classic neural networks, is because of parameter sharing. It leverages the idea that important features are the same across the whole image and a single filter can be used at every position \cite{Goodfellow}. This is a valid assumption for rectilinear images but not for LiDAR data. The representation of an object is different $5$m from the sensor compared to $20$m from the sensor. The amount of points reduces and the distance between different points increases. Figure \ref{pointcloudcars} shows how the representation of a car changes over distance. The amount of points drastically decreases, for an object of the same class and similar size, while the distance between points increases.

\begin{figure}[ht!]
\begin{subfigure}[t]{0.48\linewidth}
\includegraphics[width=\linewidth]{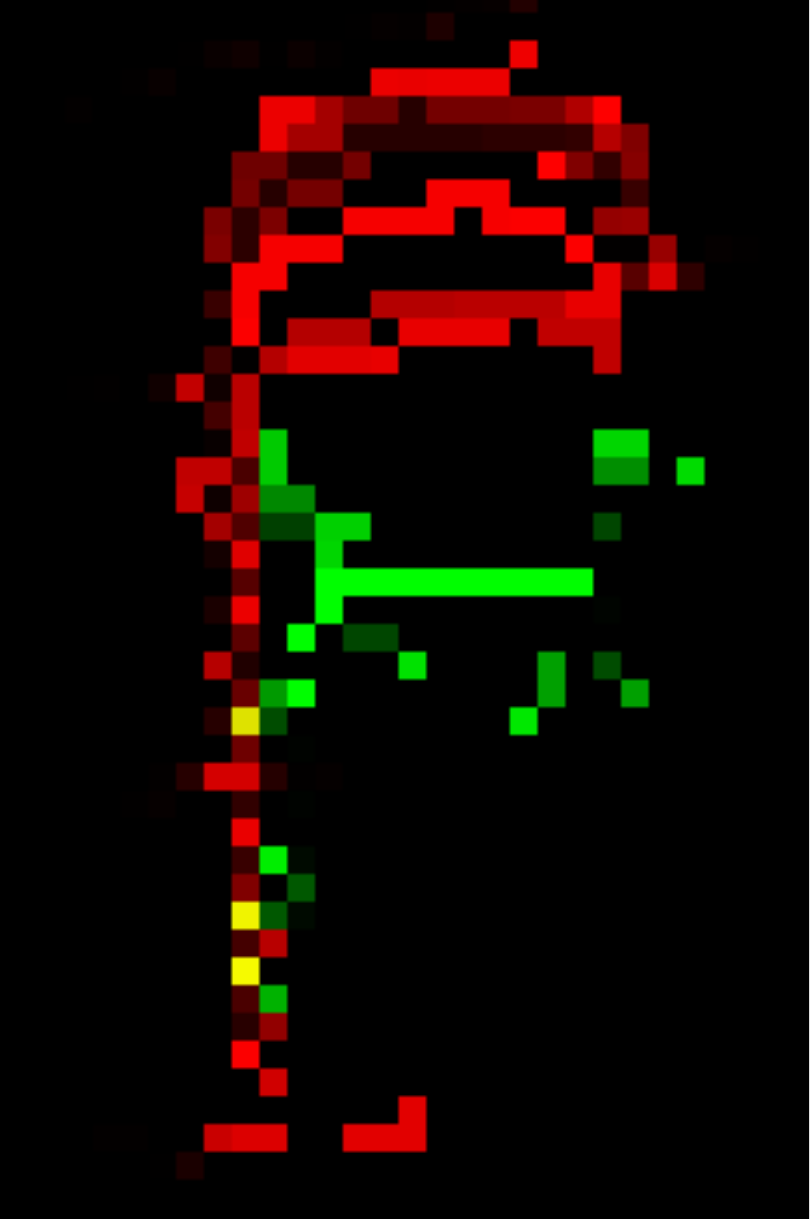}
\label{}
\end{subfigure} 
\hfill
\begin{subfigure}[t]{0.48\linewidth}
\includegraphics[width=\linewidth]{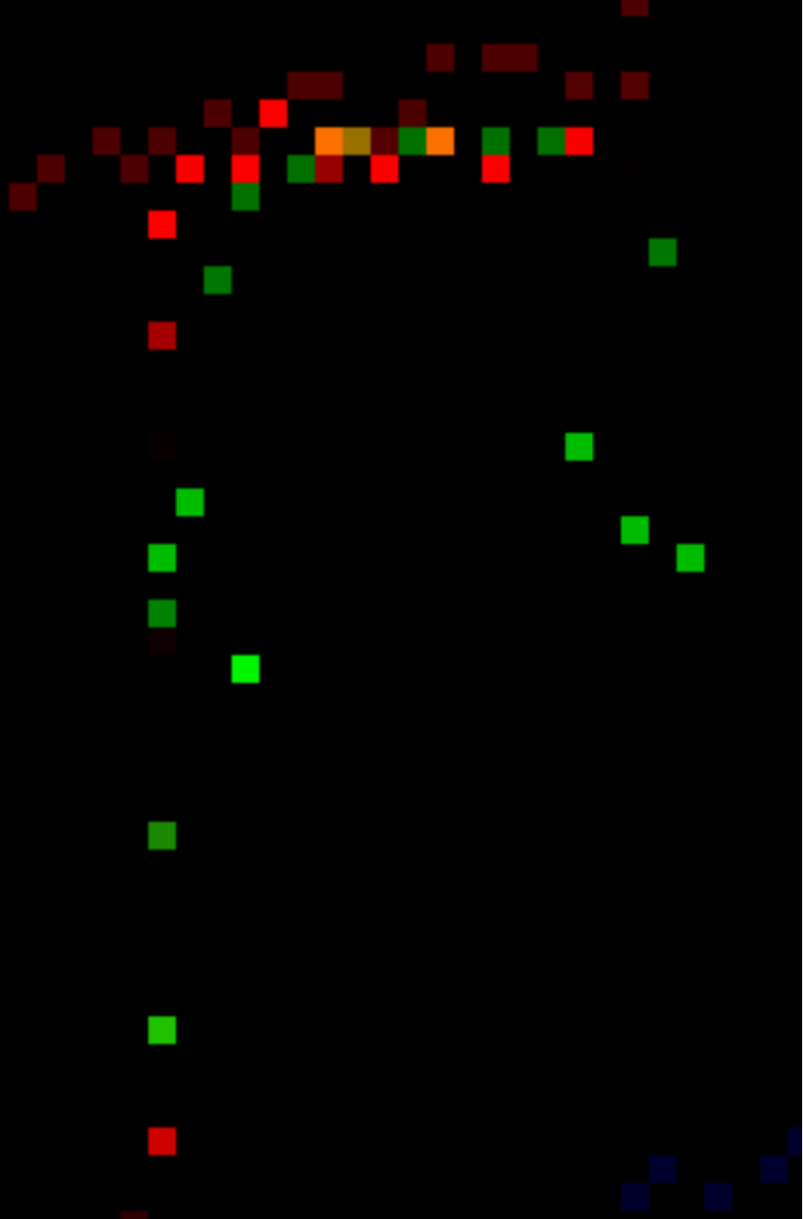}
\label{}
\end{subfigure} 
\hfill
\caption{BEV representation of two cars in the same KITTI frame. The left car is located at 20 meters from the LiDAR and the right car is located 53 meters from the LiDAR. It can be seen that the amount of points decreases over distance and gaps start appearing between adjacent pixels.}
\label{kittiimage}
\end{figure}

The sparsity and divergence of the beams over time can also be observed in BEV images. Figure \ref{kittiimage} shows the gaps that appear in the side of a car. This is not because there is not part of the car at that pixel but because the beams have diverged so far that it is simply not possible to cover all pixels with the beams. The farther away from the LiDAR, the larger the gaps. Note that these gaps in longitudinal and lateral dimension are much smaller compared to the divergence in the height dimension. The rapid divergence in the height dimension makes it difficult to detect objects that only cover a small area such as pedestrians. The green arrow in Figure \ref{pointcloudcars} represents the height dimension and shows a bigger distance than the red arrow which relates to the longitudinal and lateral dimensions. This is confirmed by looking at the specifications of the LiDAR used for the KITTI dataset, which state that the angular resolution in longitudinal and lateral range is $0.09$ degrees and the vertical resolution is $0.4$ degrees \cite{velodyne}.
The fourth car in Figure \ref{pointcloudcars} is barely distinguishable at a distance of $43$ meters. The BEV range is $70$ by $70$ meters so considering that range, $43$ meters is not even near the edge of where cars should be detected for the KITTI evaluation. 
The difficulties for far-away objects in the KITTI dataset were quite recently addressed by Wang et al. \cite{wang2019range}. They used adaptive layers to enhance the performance on distant objects and were able to improve SECOND \cite{second} and VoxelNet \cite{voxelnet} with roughly $1.5\%$ average precision for easy, moderate and hard categories. While the idea here is similar, our work does not need an adversarial approach.

\subsubsection{Combined network}

With the knowledge that features are not consistent over the range of the point cloud, the network can be adjusted accordingly. 
Multiple instances of the baseline network are used and trained on different subsets of the training data. The first network is only trained on the objects that are close by, while an identical second network is trained on objects far away. The results of the separate regions are combined and then compared to a baseline network that is trained on the full range.
To decide if objects should fall in the category "close by" or "far away" a distance threshold has to be established. From figure \ref{combinednetwork} it can be seen that a radial distance from the LiDAR sensor is used.

For the inside range, the point cloud is converted to a BEV image in the same manner as before, but now all the pixel values outside this threshold range are set to zero. This is done similarly for the outside range but the inside set of pixels is set to zero as displayed in Figure \ref{combinednetwork}. The point cloud is converted to two BEV images. Each one contains part of the point cloud and part of the objects.
If the distance to the center of an object is smaller than the threshold range it belongs to the inside network and if it is larger it belongs to the outside network.

When objects are excluded based on their center, it could happen that objects are partially cut out of the image and fed to the network. This should not occur since it will devalue the training data. To avoid this, an overlap region is introduced. This region is of the point cloud is used both for the inside and outside range networks.

\section{\uppercase{Implementation details}}

\subsection{Network settings}
All tests are done on a Faster R-CNN network with a rotational branch inspired by \cite{detectionteamUCas}. A learning rate of $0.0003$ with decay steps at $190k$ and $230k$ with a decay factor of $3$ for both steps is applied. The dataset, which contains $7481$ samples, is split in training, validation and testing in a $50/25/25$ fashion. 
For all networks, multiple checkpoints were evaluated and the one with the best score on the validation set was picked. The best scores occurred often between $200k$ and $250k$ steps
which means that the network were trained for roughly $50$ to $60$ epochs. All networks are trained on a single Nvidia Tesla V100 GPU.
Anchors of $45\times20$ pixels are used to detect cars which is in line with the average car size.
 
The network is optimized using stochastic gradient descent with momentum of $0.9$. Weight decay of $0.0001$ is used to prevent overfitting. 
All weights of the ResNet backbone are pre-trained on ImageNet \cite{imagenet}.
The underlying data is quite different from the LiDAR data but we observed fast convergence for the network.
A batch size of $1$ is used with batch normalization applied with a decay factor of $0.997$. Batch normalization is applied but not trained, the values from the pre-trained ImageNet weights are used for the batch normalization layer. 

Smooth L1 loss is used with thresholds of $3$ and $1$ for the RPN and head network respectively. The loss weights for the different components are found empirically. It is important to note that only one class is considered during training, which makes classification easier. Most errors occurred in the box calculation so the regression loss is twice as high as the classification loss for both RPN and head network branches.

\begin{figure}
\includegraphics[width=0.48\textwidth]{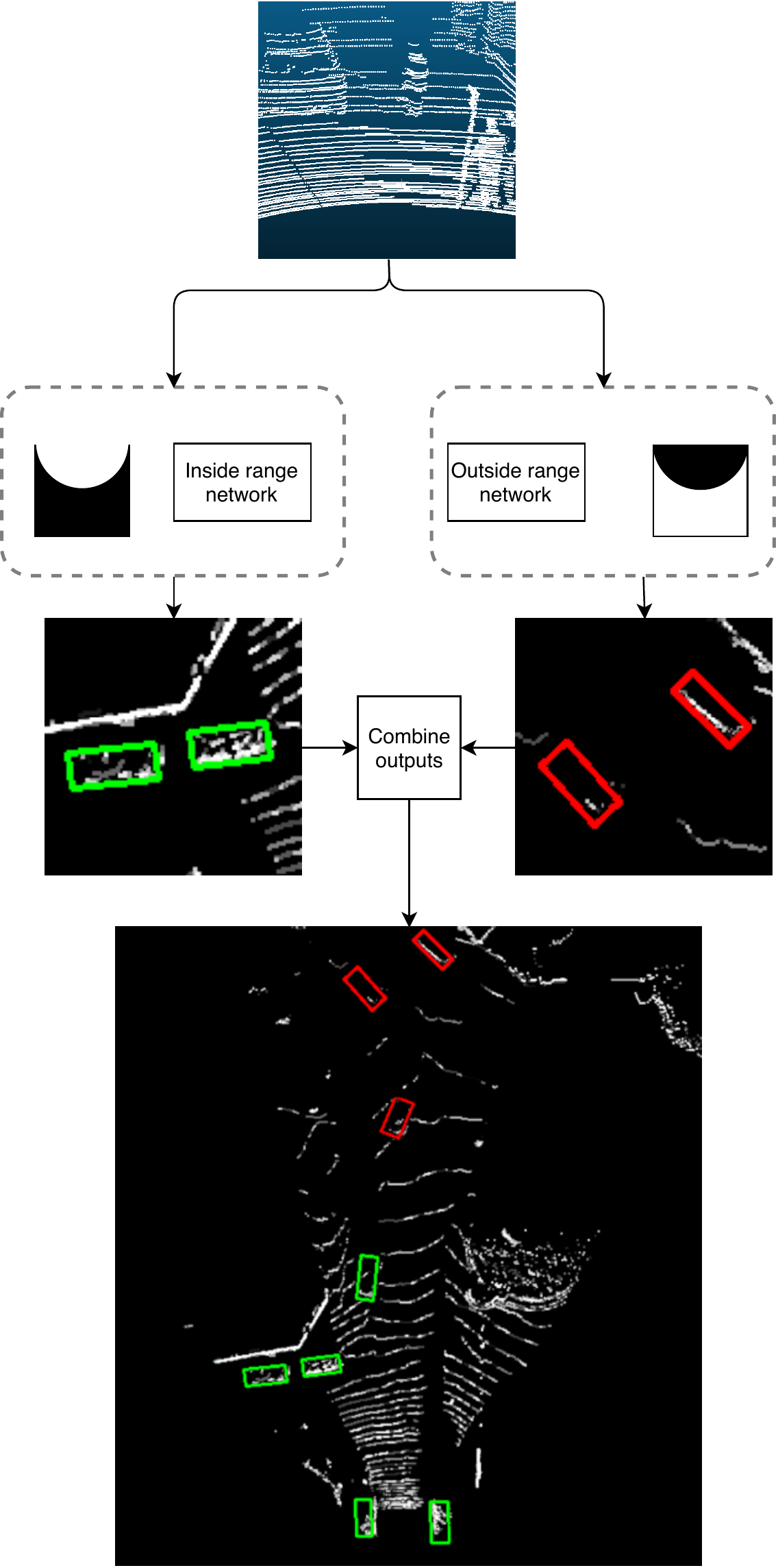}
\caption{Pipeline of the combined network architecture. Inside and outside range network only consider the white part of the visualized range next to them. These networks have the exact same configuration and are in line with Figure \ref{pipeline}. The outputs of both networks are combined to form the final output.}
\label{combinednetwork}
\end{figure}

\subsection{Data pre-processing}

Point clouds are converted to one of the four configurations described in the Methods section, displayed in Figure \ref{bevconversion}. Every pixel corresponds to $0.1\times0.1$ meter space in the point cloud. All tests are done on the KITTI dataset where only the field of view (FOV) of the front camera is annotated. All pixel values outside this FOV are set to zero which results in a region of black pixels. Some objects that lay on the border are sometimes still annotated. To generalize this situation for all images, only objects that are annotated and have at least $50\%$ of the surface in the FOV are taken into account.
No augmentations are used for the final models because they did not improve the performance of the network.

\subsection{Combined network}

The combined network uses the exact same settings described in the previous sections, but applied to  different range. One network is trained on data close to the LiDAR while the another network is trained on data far away from the LiDAR. The final output is their combined output for the respective regions. For the inside range network only boxes with the center closer than $25$m to the LiDAR are taken into account. With a $5$m overlap region between $25$m and $30$m, to make sure no labeled objects are cut in half.
For the network that trains on objects that are far away, a $30$m boundary is used where the $25-30$m region is the overlap region mentioned before. Figure \ref{combinednetwork} shows how the images are combined.
The choice of the threshold for a particular dataset affects the results of the network. We found that $25$m was a good threshold since both regions still contain enough training samples.

For inference a more straight forward division is used.
The KITTI evaluation considers a range of [$-35$m, $35$m] laterally and [$0$m, $70$m] longitudinally. We simply divided this region in half to get a close by and far away range where we evaluate the baseline and combined network on.
All objects in the top half of the image bounded by [$-35$m, $35$m] laterally and [$0$m, $35$m] longitudinally will be detected by the inside network. The outside network detects objects in a space bounded by [$-35$m, $35$m] laterally and [$35$m, $70$m]. The results of both networks are then merged together. The full pipeline is displayed in Figure \ref{combinednetwork}.

\section{\uppercase{Results}}

\noindent
All results are based on the KITTI benchmark which calculates average precision (AP) scores for three different categories, respectively easy, moderate and hard. AP is an often used object detection metric. In KITTI, $41$ equally spaced recall points are used where the precision at each point is calculated.
Precision is evaluated at different recall points and combined according to the following equation:

\begin{equation}
AP_\textrm{kitti} = \frac{1}{41}\int_{0}^{1} p(r) dr
\end{equation}

where $r$ is a set of $41$ values linearly spaced between $[0, 1]$ where each value is a specific recall value. The integral of all recall values divided by $41$ is the AP score. 

Different true positive thresholds are considered. The official KITTI has a $70\%$ intersection over union (IoU) threshold for the car class but we also report scores for $50\%$ overlap as many other works analyse. $70\%$ percent is quite challenging to achieve with rotational bounding boxes around objects that are often occluded or contain few points.

%0.7 AP
\begin{table*}[t]\centering
\resizebox{0.98\linewidth}{!}{%
  \begin{tabular}{|c||c|c|c||c|c|c||c|c|c|}
    \hline
    & \multicolumn{3}{c||}{0-35m range} & \multicolumn{3}{c||}{35-70m range}  & \multicolumn{3}{c|}{0-70m range} \\    
    \hline
Method (Threshold) & Easy & Moderate & Hard & Easy & Moderate & Hard & Easy & Moderate & Hard \\ \hline
Baseline (0.7)  & 78.8 & 82.0     & 75.4 & -    & 43.7     & 37.0 & 78.5  &  73.0     & 66.9 \\ 
Combined Network (0.7) & 81.5 & 83.4     & 76.5 & -    & 46.5    & 42.0 & 81.2  & 74.7     & 68.9  \\ 
Difference   & \textbf{2.7} & \textbf{1.4} & \textbf{1.1} & \textbf{-} & \textbf{2.8} & \textbf{5.0} & \textbf{2.7} & \textbf{1.7} & \textbf{2.0} \\ \hline
Baseline (0.5)  & 89.3 & 89.7    & 89.2 & -    & 60.9     & 53.0 & 89.0  &    82.9   & 81.2 \\ 
Combined Network (0.5) & 89.8 & 89.9    & 89.5 & -   & 75.1    & 67.7 & 89.5  & 86.4      & 84.7  \\ 
Difference   & \textbf{0.5} & \textbf{0.2} & \textbf{0.3} & \textbf{-} & \textbf{14.2} & \textbf{14.7} & \textbf{0.5} & \textbf{3.5} & \textbf{3.5} \\ \hline
  \end{tabular}}
  \caption{Baseline network compared to the combined network for 50\% and 70\% IoU threshold. The average precision for the 0-70 meter range is the weighted mean average precision of the 0-35 range and the 35-70 range. The reason for this is that the confidence scores of the combined network do not match. The confidence scores of the objects far away are too high because that network has never seen objects close by. This difference in confidence score between the two networks influences the AP calculations, so a mean average precision of the two ranges is used.}
  \label{combinedrange07}
\end{table*}

\subsection{Feature analysis}
Multiple different input features and their impact on results are tested next.
In the methods section, the considered four different input configurations are explained. 
Table \ref{inputfeatures} shows the results on the KITTI benchmark for the different pre-processing methods displayed in Figure \ref{bevconversion}.
It can be seen that the first three methods, maximum heights, binary and height/intensity/density only differ by a small margin. Only the approach where 9 channels are used performs significantly worse. 

\begin{table}\centering
\resizebox{0.98\linewidth}{!}{%
  \begin{tabular}{|c||c|c|c|}
    \hline
    
    \hline
Features & Easy & Moderate & Hard \\ \hline
max height   & 79.5 & 73.1 & 66.6  \\ 
height, intensity, density  & 79.2 & 73.1 & 67.0  \\ 
binary   & 79.4 & 72.9 & 66.4 \\
multichannel  height vox. & 76.0 & 65.8 & 65.0  \\  \hline
  \end{tabular}}
  \caption{Results of different input feature configurations.}
\label{inputfeatures}
\end{table}

Overall the networks seem to learn more the structure of the objects and not the absolute values stored in the channels. This does not come as a complete surprise since a ResNet feature extractor with batch normalization is used. The information of the absolute values is "lost" quite quickly in the network because of the normalization. The filters in the first layers search more for derivative-based features such as edges and corners. From that perspective, the differences are not so big. With that being said, the input values are still important for the performance of the network. Solely using intensity has a much worse performance than only using the max height value \cite{birdnet}. Another factor could be the pre-trained weights. That allows for fast training results but may limit the overall performance of the network. This could be a problem especially for the network with nine channel inputs.

These results highlight a limitation of 2D detectors processing on point clouds and a possible reason for the gap between them and methods directly using the point cloud.
For LiDAR data these absolute values are of big importance, since they can give information about classes directly. An object is very unlikely to be a car if the highest values are only $1$ meter high even though the top view almost perfectly corresponds. It is possible that methods that use the point cloud directly are able to leverage this information more than 2D detectors. These networks also use batch normalization but are able to compute more complete features that help boost their scores. Relying too heavily on these input values could be dangerous when considering different classes, noise, sensor movement, etc. With the rise of new datasets, it will hopefully become clear how robust these methods are.

\subsection{Range analysis}
Table \ref{combinedrange07} shows the results of the baseline network and the combined network on a $70\%$ IoU threshold and a $50\%$ threshold. Easy category is not considered for 35-70m range analysis, due to the lack of significant amount of cars. The combined network outperforms the baseline network for all categories. The inside and outside networks are trained on subsets of the total data and are still able to perform better than the baseline network. Not sharing the weights results in a better performing network, because objects in different ranges do not have to share the same feature extractors.

The results of $50\%$ IoU compared to the ones with  $70\%$  show that even if the networks are able to detect the cars, in many cases the precision of the bounding boxes hurts the results significantly. Cars that are closer to the LiDAR contain more points and regressing their boxes is easier in these situations. If only a $50\%$ threshold is used, the evaluation is more forgiving for errors in rotation because they have the largest influence on the overlap. For the 35-70m range the improvements of the combined network are remarkable. It seems that the difference in features over the distance hurts the ability to detect objects, when the model is trained with full range data. The same happens with the ability to detect the orientation of the objects accurately. In automated driving this is important since the orientation is key for further processing steps such as estimating heading angles and applying tracking.

\begin{figure*}[ht!]

\begin{subfigure}[t]{\linewidth}

\includegraphics[width=0.45\linewidth, height=6cm]{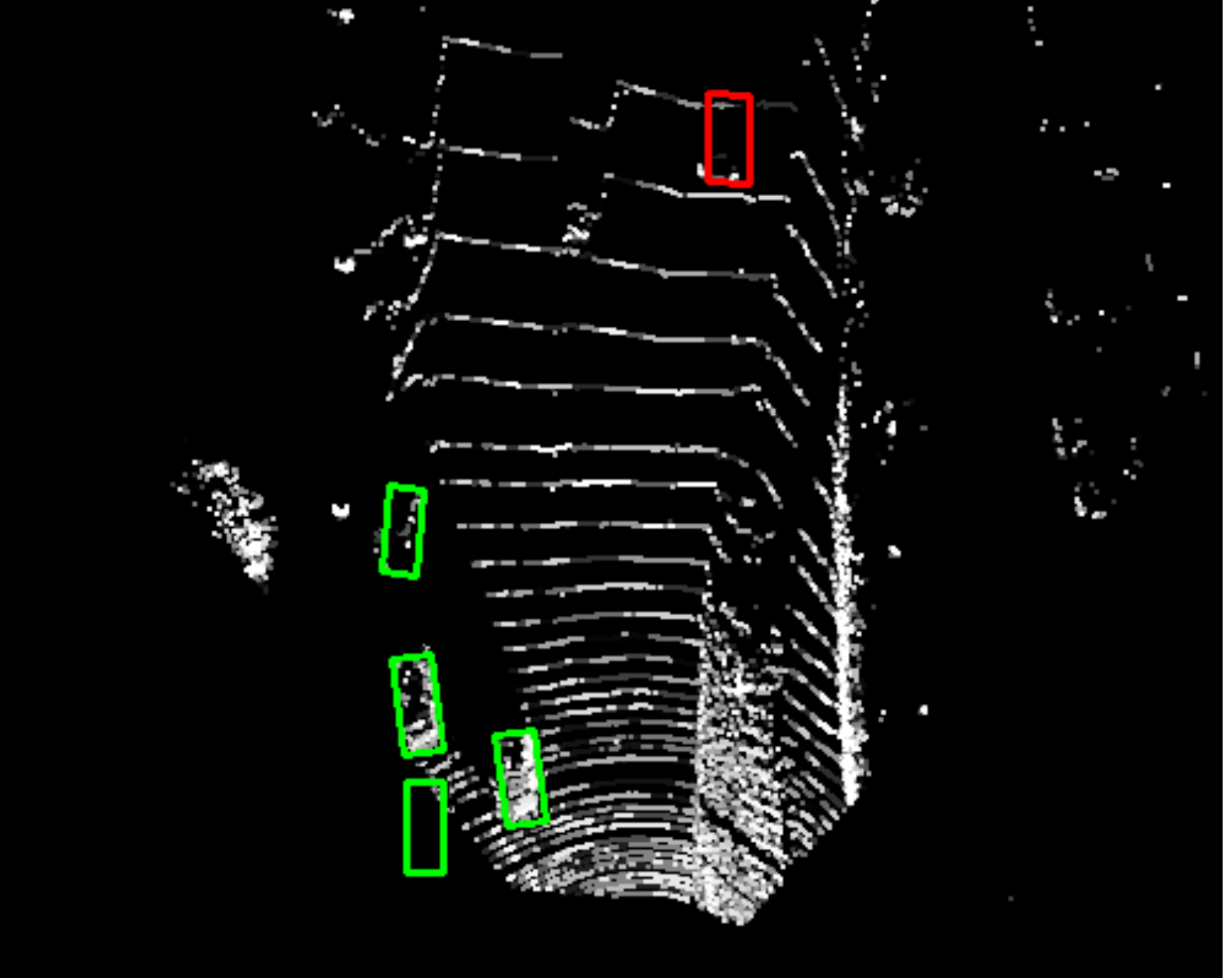}\hfill
\includegraphics[width=0.45\linewidth, height=6cm]{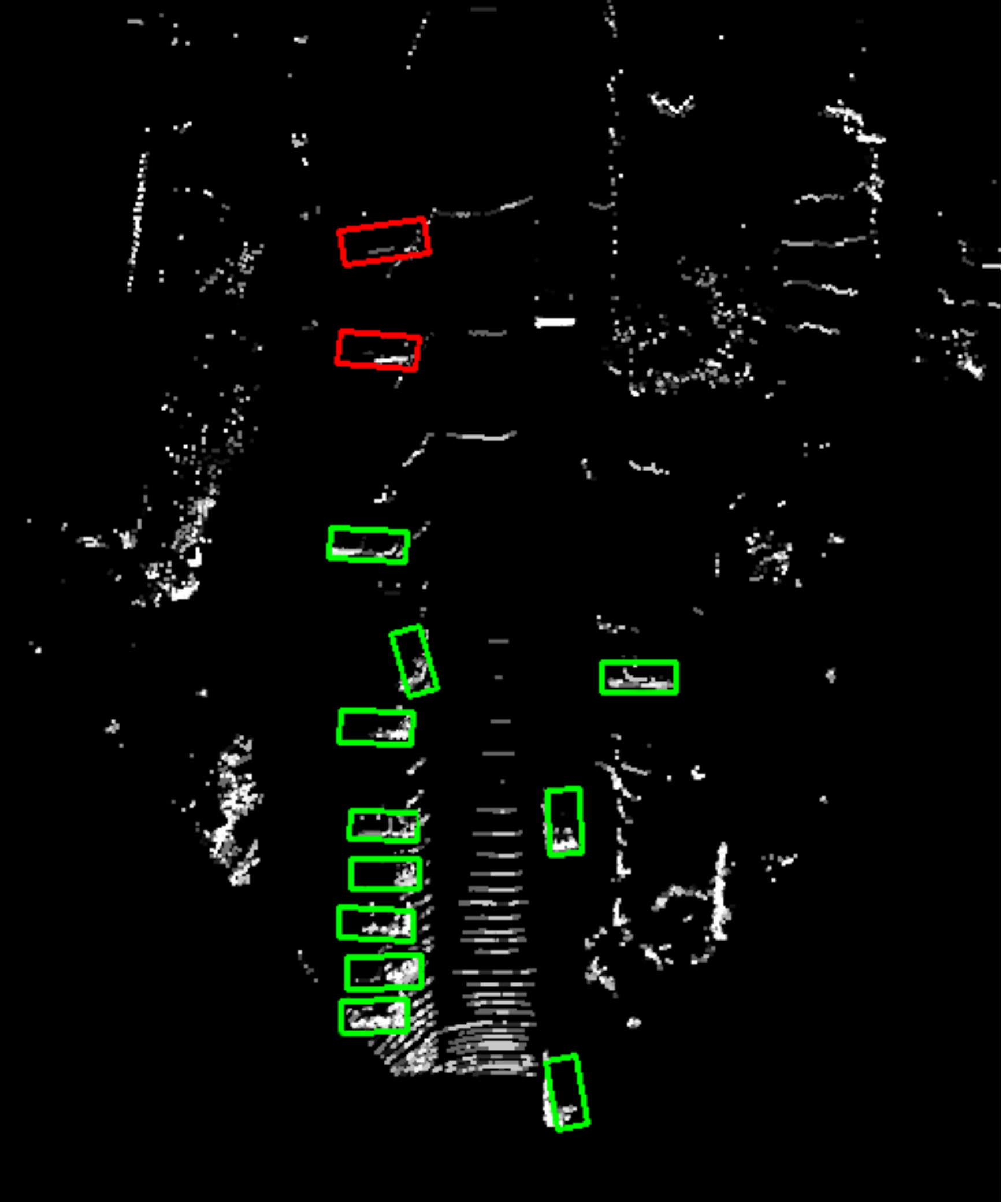}

\includegraphics[width=0.45\linewidth, height=3cm]{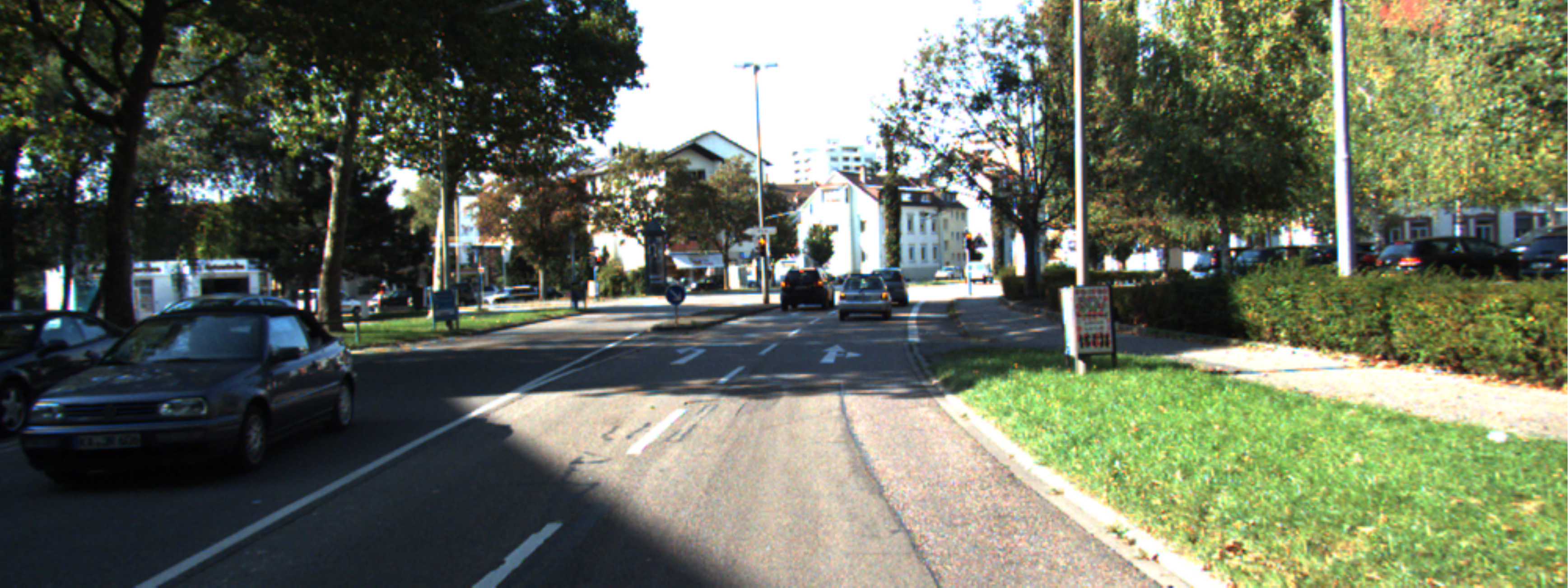}\hfill
\includegraphics[width=0.45\linewidth, height=3cm]{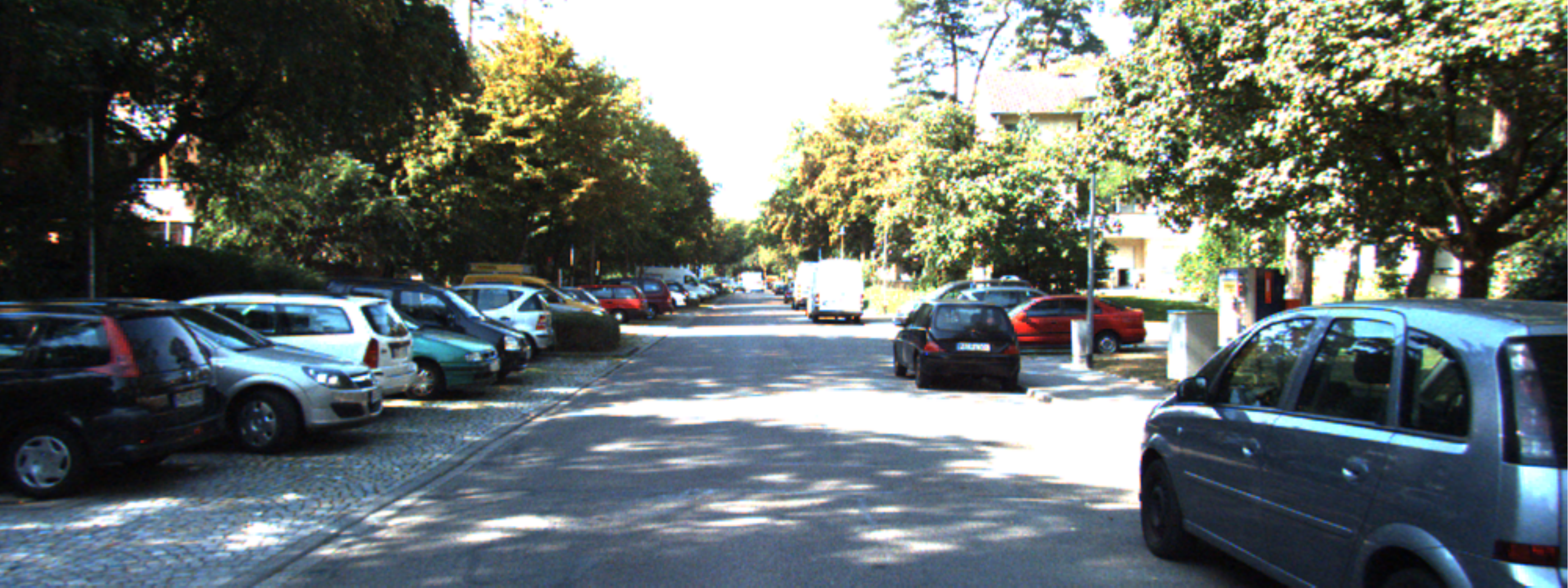}
\end{subfigure}\\

\caption{Detection results on KITTI validation set. The upper row in each image is the BEV representation with the detected cars. The other are the camera images corresponding to those BEVs. Green boxes are estimated by the inside network and red boxes by the outside network.}
\label{finalimages}
\end{figure*}

Figure \ref{finalimages} shows qualitative results of the combined network. The green boxes are from the inside network and the red boxes from the outside network. The BEV image is adjusted for better visualization by smoothing the images and maximizing the input values.
In the left image, it can be appreciated that the closest bounding box to the LiDAR is a correct detection, although almost all of the pixels that represent the car are black. The algorithm is able to detect it correctly based on the small corner that is still present.
The right figure shows how the network performs all over the range.

\section{\uppercase{Discussion}}
\label{sec:conclusion}

\noindent A 3D object detection algorithm is implemented and trained on the KITTI dataset, that outperforms many recent BEV based networks \cite{complexer-yolo,birdnet,lasernet} and gets similar results to other networks \cite{Pixor,m3vd}.
This paper shows how convolutional neural networks used for natural images are based on assumptions that do not transfer well to point cloud data. This concept can be used to increase the performance of many networks on the KITTI benchmark whether they are processing BEV images or point clouds directly as was also shown by \cite{wang2019range}. Furthermore it looks into the advantages and limitations of BEV approaches. The main advantage is robustness since it does not rely on the specific values of the object, but mostly on the shape. This could also be thought as a limitation since these specific values could be directly used for classification in LiDAR data. However, these values can vary between LiDAR models and are more sensitive to possible noise. Data quantity that needs to be processed is another thing to take into account, as only an image needs to be processed instead of an entire point cloud.

\section{\uppercase{Conclusions}}

\noindent This work provides insights in how LiDAR data differs from natural images, which features are important for detecting cars and how a neural network architecture for 3D object detection can be adjusted to take into account the changing object features over the distance.
The first tests check the effect of different handcrafted input features on the performance. Point clouds are converted to BEV images that are fed to a two stage detector. Different input configurations do not vary much in performance which can be attributed to the fact that the filters look for features such as edges and do not rely as much on the exact values in the BEV. These exact values are of more relevance for LiDAR data compared to camera images because they can be linked directly to certain classes. This is different from approaches that use the raw point clouds directly as input and might explain why they perform overall better on the KITTI benchmark. Nevertheless, relying on those values may be a problem when dealing with sensor noise or model differences. With the rise of new datasets, it will hopefully become clear how robust these methods are compared to BEV based approaches.

In addition, this work visualizes how the distance to the sensor influences the objects representation in the point cloud.
Most convolutional neural network rely on the assumption that features are consistent over the full range of the image. This allows for one filter to be used to extract features from the entire feature map. For LiDAR data this is not the case which is shown by analyzing point clouds and objects at various distances. This observation is used to change the detection pipeline and have a separate detector for objects in the 0-35 meter range and another detector for objects in the 35-70 meter range. These changes lead to improvements, most notably of 2.7\% AP on the 0-35 meter range for easy category and 5.0\% AP on the 35-70 meter range for hard category, using a 70\% IoU threshold.

\bibliographystyle{IEEEtran}
{\small
\bibliography{main}}

\end{document}